\documentclass[letterpaper, 10 pt, conference]{ieeeconf}  
\usepackage[latin1]{inputenc}
\usepackage[T1]{fontenc}
\usepackage[english]{babel}
\usepackage{amsmath}
\usepackage{amssymb,amsfonts,textcomp}
\usepackage{color}
\usepackage{multicol}
\usepackage{array}
\usepackage{hhline}
\usepackage{graphicx}
\usepackage{gensymb}
\usepackage{hyperref}
\usepackage{booktabs}
\usepackage{newunicodechar} 
\usepackage{amsmath}
\usepackage{mwe}
\usepackage{authblk}
\usepackage[linesnumbered,ruled]{algorithm2e}
\usepackage{caption}
\usepackage{subcaption}
\hypersetup{pdftex, colorlinks=true, linkcolor=blue, citecolor=blue, filecolor=blue, urlcolor=blue, pdftitle=, pdfauthor=vaibhav, pdfsubject=, pdfkeywords=}

\IEEEoverridecommandlockouts                              
\title{\LARGE \bf
Dynamic Motion Planning for Aerial Surveillance on a Fixed-Wing UAV}

\author{Vaibhav Darbari\qquad Saksham Gupta\qquad Om Prakash Verma
}
\affil{Department of Computer Science and Engineering\\Delhi Technological University\\ Rohini, Delhi 110042, India\\vaibhavdarbari@gmail.com, sakshamgupta\_2k14@dtu.ac.in, opverma@dce.ac.in}

\begin{document}

\maketitle
\thispagestyle{empty}
\pagestyle{empty}

\begin{abstract}
We present an efficient path planning algorithm for an Unmanned Aerial Vehicle surveying a cluttered urban landscape. A special emphasis is on maximizing area surveyed while adhering to constraints of the UAV and partially known and updating environment. A Voronoi bias is introduced in the probabilistic roadmap building phase to identify certain critical milestones for maximal surveillance of the search space. A kinematically feasible but coarse tour connecting these milestones is generated by the global path planner. A local path planner then generates smooth motion primitives between consecutive nodes of the global path based on UAV as a Dubins vehicle and taking into account any impending obstacles. A Markov Decision Process (MDP) models the control policy for the UAV and determines the optimal action to be undertaken for evading the obstacles in the vicinity with minimal deviation from current path. The efficacy of the proposed algorithm is evaluated in an updating simulation environment with dynamic and static obstacles.
\end{abstract}

\section{INTRODUCTION}
In the last decade, Unmanned Aerial Vehicles (UAVs) have gained importance in a myriad of military
and civilian applications. They are especially useful in missions pertaining to surveillance, rescue and exploration. In these tasks UAVs are often required to operate in
unfavourable conditions where a complete description of operating environment may not be available.A fully autonomous UAV eliminate the need to plan the mission beforehand in an unknown , unpredictable environment, it is capable of handling any new situation which is thrown upon it spontaneously.
Consider the following scenario:An urban setting where a UAV is released in an unknown airspace for the purpose of surveillance. The aim of the UAV is to maximize mapped area with constraints such as minimum turning radius and avoiding collision with other planes/UAVs operating in vicinity, buildings , other dynamic and unknown obstacles.In such a challenging scenario fully autonomous path planning plays a vital role in ensuring success of the mission specially for fixed-wing UAVs which lack hovering capability and hence need to maintain a minimum airspeed to maintain lift.
\par{
Several techniques have been proposed for 3D path planning like Rapidly-exploring Random Graph\cite{c1},Visibility Graphs\cite{c2},PRM\cite{c3},RRT*\cite{c4},Artificial Potential Field. All these techniques though effective,are computationally expensive and need to be recomputed as soon as the environment is changed.The above mentioned techniques suffer the curse of dimensionality and a combinatorial explosion takes place with increase in search space.}
\par{
This paper proposes a 3D Dynamic path planning approach which overcome the aforementioned limitations and facilitates conduction of fully autonomous missions.
The rest of the paper is organized as follows In Section II:Scenario describing the environment,obstacles and coordinate space is presented.Section III comprises of proposed approach giving details of global and local path generation along with the decision process for optimal control policy generation.The simulation results and analysis for the proposed approach are presented in Section IV.We conclude with final remarks and discuss future works in Section V. }

\section{Scenario}
A fixed-wing autonomous UAV is to conduct surveillance of a pre-specified area(Search Space) within a given Geofence.Considering the given scenario, the aim of our approach is to:
\begin{itemize}
	\item Maximize the area surveyed in minimum flight time. 
	\item Avoid collision with any obstacle, be it stationary or dynamic.
\end{itemize}
 The UAV has a constraint on its turning radius , climb-rate and minimum airspeed.The path planning problem is finding the optimal path in the search space given the constraints on UAV and environment.
\subsection{Environment}
We assume that the UAV is GPS enabled and carries some specialized sensors capable of accurately detecting the location of the obstacles.
The search space is assumed to be partially known for the PRM and the approach is designed keeping in mind that any kind of obstacle can turn up at any coordinates on any given time.
\subsubsection{Static Obstacles}
Static or Stationary obstacles are assumed to be a solid cylinder. The coordinates of the centre of the cylinders along with its radius and height is updated whenever there is a change in the environment.
\subsubsection{Dynamic Obstacles}   
Dynamic or Moving obstacles are assumed to be solid spheres. Similar to static obstacles the coordinates of sphere's centre, radius and velocity are synced with the changes in the environment.The trajectory of the dynamic obstacles are randomly selected from a pool of precomputed paths so as to mimic real life aerial vehicles that may operate in the vicinity of the UAV.  

\subsection{Coordinates Transformation}
\subsubsection{WGS 84 to ECEF coordinates}
Most navigation systems are based on World Geodetic System including GPS. WGS is modelled after the standard coordinate system for Earth, a standard ellipsoid reference surface for raw altitude data, and a gravitational equipotential surface that defines the nominal sea level. 
We need to convert to a Local Tangent Plane Coordinate System for the ease of mathematical computation  so as to make the problem retractable. The order of approximation involved in the conversion is of the magnitude of $10^{-9}$ for the small distances involved for the scenario of our paper. Geodetic coordinates(latitude $\lambda$, longitude $\varphi$, height \textit{h}) can be converted to ECEF coordinates using the equations:

\begin{equation}
	X = (N (\lambda) + h)cos\lambda  cos\varphi
\end{equation}	
\begin{equation}
	Y = (N (\lambda) + h)cos\lambda  sin\varphi
\end{equation}
\begin{equation}
	Z = (N (\lambda)(1-e\textsuperscript{2}) + h)sin\lambda	
\end{equation}
	
where
\begin{equation}
	N(\lambda ) = \frac{a}{\sqrt{1-e^{2}sin^{2}\lambda  } }
\end{equation}
here $a$ and $e$ are semi-major axis and the first numerical eccentricity of the ellipsoid respectively. The values of which can be found in the definition of WGS 84\cite{c5}.
\subsubsection{ECEF to WGS 84}
We need to transform the ECEF coordinates back to geodetic coordinate system so as to facilitate the waypoint planning on the on-board autopilot system.

\begin{equation}
	\lambda = \arctan\frac{Y}{X}
\end{equation}
\begin{equation}
	\varphi = \arctan\frac{Z + e'^{2}bsin^{3}\omega}{p-e^{2}acos^{3}\omega }
\end{equation}
\begin{equation}
             h = \frac{p}{cos\varphi} - N
\end{equation}  

where auxiliary values are:
\begin{equation}
	p = \sqrt{X^{2}+Y^{2}}
\end{equation}
\begin{equation}
	\omega = \arctan\frac{Za}{pb}
\end{equation}
	
\section{Proposed Approach}
\subsection{Overview}
\par{Two phase motion\cite{c6} planning is a widely adopted strategy for reducing the planning complexities of high-order dynamic systems\cite{c7}\cite{c8}. At the first stage, a coarse but feasible path is generated taking into consideration the kinematics of the system and the known static elements of the environment. System design constraints and dynamic elements in the environment may be ignored in this phase. The primary objective of this phase is to confine the search space to the vicinity of the generated path and thus reduce the complexity of the planning task. At the second stage, a smoother more finely attuned path is generated in response to changes in the environment and dynamics of the system. The two stages compensate each other in terms of planning complexity, risk associated with the path and type of obstacles handled. Two phase planning combines the merits of both the methods and facilitates real-time operation in a cluttered environment. }
\begin{figure*}[t!]
	\centering
	\includegraphics[width=0.75\linewidth]{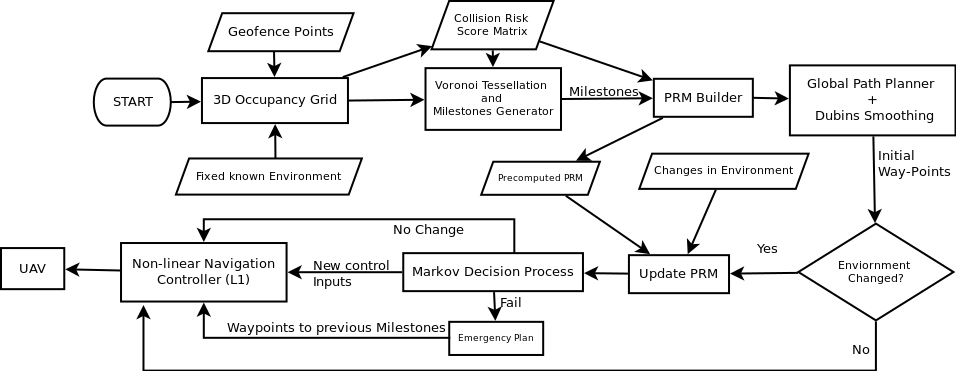}
	\caption{Overview of the proposed architecture}
	\label{arch}
\end{figure*}
\par{The major components of our system along with the flow of control are represented in system architecture shown in the Fig.\ref{arch}. Global planner works in a 3D occupancy grid with boundaries demarcated by the geofence. The global planner takes starting location, geofence and known static elements of environment as input and generates a tour of search space with objective of maximizing area coverage while minimizing collision risk. The output of the global planner is a sequence of waypoints representing the tour. Certain waypoints in the search space are marked as milestones to avoid being trapped in local minima by virtue of heuristic search algorithm and to ensure maximum area coverage (details in later sections).Local planner works alongside the global planner and caters to moving obstacles and unforeseen stationary obstacles. Local planner checks for potential collision on the current path and assesses the risk associated with it. In case current path is deemed risky, the path is modified with a sequence of dynamic waypoints to avert the danger. The trajectory of the modified path is based on smooth Dubin curves to ensure a dynamically feasible and steerable path.
A decision process determines the type of path adjustment to be undertaken by the global or local planner and accordingly waypoints maybe modified.
If both original and modified paths are deemed risky, the current milestone is marked unsafe and local planner computes a route to the nearest unvisited safe milestone. In absence of such a milestone UAV is guided to last known safe milestone. When each milestone has been visited the UAV returns to the starting location and lands. A detailed explanation for each of the phases follows in the subsequent sections. 
}
\subsection{Global Planner}

As mentioned earlier, the objective of global planner is to compute a kinematically feasible tour of the search space which maximizes area covered and minimizes collision risk. Initially, a discretised 3D occupancy grid is computed from the search space with each cell value indicating the risk associated with it. The grid is then sampled randomly for safe cells. A probabilistic road map (PRM) is built using the computed samples. The samples are clustered into safe regions and the centroid of every safe region is considered a milestone. Starting from the initial location, a greedy approach is used to form a tour going through all the milestones. The optimal path between individual milestones is determined by A* algorithm. The path so formed is bound to be kinematically viable as for each node in the path, only successors following the constraints are considered for path expansion.

\subsubsection{Building Occupancy Grid}

The search space within the confines of the geofence is represented as a discretised 3D grid. Each cell of the grid is associated with a collision risk involved with the cell. 
A global occupancy score for a cell is determined from the obstacle field distribution in the search space. Both static and dynamic obstacles contribute to this obstacle field. The global occupancy score for a cell is expressed as

\begin{equation}
S_{\mathit{Global}}^{\mathit{(x,y,z)}}=\left\{\begin{matrix}1,\mathit{if}\mathit{the}\mathit{cell}\mathit{space}\mathit{overlaps}\mathit{with}\mathit{any}\mathit{obstacle}\\\frac{\sum
	_{i=1}^k\frac 1{D_i^{\mathit{(x,y,z)}}-R_i}}
k,\mathit{if}\mathit{cell}\mathit{lies}\mathit{outside}\mathit{obstacle}\end{matrix}\right.
\end{equation}

where $k$ is the count of known obstacles, $D^{(x,y,z)}_i$is the Euclidian distance of the cell centred at (x,y,z) from the $i_{th}$ obstacle and $R_i$ is the radius of the  $i_{th}$  obstacle. This score provides an accurate estimation to the proximity of the obstacle but leads to a coarse distribution of values in the occupancy grid around the obstacles. Thus a local score is also considered alongside the global score to facilitate a evener distribution of values in the grid. 
\begin{equation}
S^{(x,y,z)}_{Local}=\sum^{x+1}_{i=x-1}{\sum^{y+1}_{j=y-1}{\sum^{z+1}_{k=z-1}{S^{(i,j,k)}_{Global}}}} 
\end{equation} 
The local score, denoted as $S^{(x,y,z)}_{Local}$ for a cell centred at (x,y,z), is given as the mean of values of all the adjacent cells in the grid.The collision risk for the cell is taken as the maximum of its local and global occupancy score.
\begin{equation}
	C_{Risk}=Max(S^{(x,y,z)}_{Global},S^{(x,y,z)}_{Local})
\end{equation}
\begin{figure}[h!]
	\includegraphics[width=0.95\linewidth]{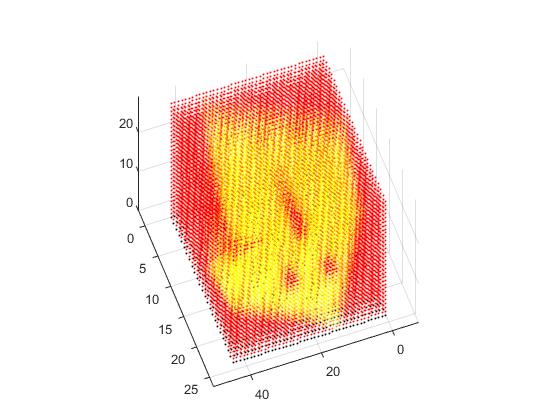}
	\caption{Occupancy Grid for three obstacles inside geofence.The area inside,around the obstacle and in the vicinity of geofence receive a high $C_{Risk}$ according to scoring policy and is thus shown in red in the figure.Safer area are marked with shades of yellow.}
	\label{probmat}
\end{figure}
\subsubsection{Voronoi Tessellation of search space}\label{voronoi}
 To ensure maximum coverage of search space for surveillance, it is necessary to ensure that the path planning algorithm does not remain confined to any one region as a result of being trapped in local minima which have been a problem with many planners including potential field-based methods. Thus there exists a need for division of target space (free space) into disjoint sub regions which must each be covered to facilitate maximum coverage.Voronoi tessellation of search space can provide the required partitions given the seed points. For a typical UAV mission, the number of seed points can be estimated to be roughly equal to the number of identified critical tasks.  For a discretized configuration space as considered, voronoi tessellation can be achieved by means of K-Medoids Clustering\cite{c8} with $K$ being set to the number of milestones.The objective function minimized during the iterations is given in Eq.\ref{eq1}. 
 \begin{equation}\label{eq1}
 J = \sum_{j=1}^{K}\sum_{i\epsilon C^{j}}d(x^{(i)},z^{(j)})
 \end{equation}
 where $K$ is the number of milestones,$z^{(j)}$ is medoid of the $j^{th}$ cluster, $x$ is a sample data point and $C^{j}$ is the  $j^{th}$ cluster. 
 \par{Only a randomly chosen subset of safe points in target space are considered for clustering due to the high runtime complexity associated with the procedure.A sample cell is considered safe if the associated collision risk score is less than the collision risk threshold $'\delta'$.An adequate number of sample points are chosen to be a good indication of the overall
 distribution of large target spaces within the search space.}
\par{For a given scenario a good heuristic for estimating the number of milestones can be taken as
	}
\begin{equation}\label{fov}
	K = \frac{\Delta_{geofence}}{\pi h_{UAV}^{2}\tan^{2}{\frac{\theta_{fov}}{2}}}
\end{equation}
where $\Delta_{geofence}$ is the area of geofence, $h_{UAV}$ is the anticipated cruise altitude for the UAV for in the mission and $\theta_{fov}$ is the field of view of the onboard camera.
 \begin{figure}[h!]
 	\includegraphics[width=0.75\linewidth]{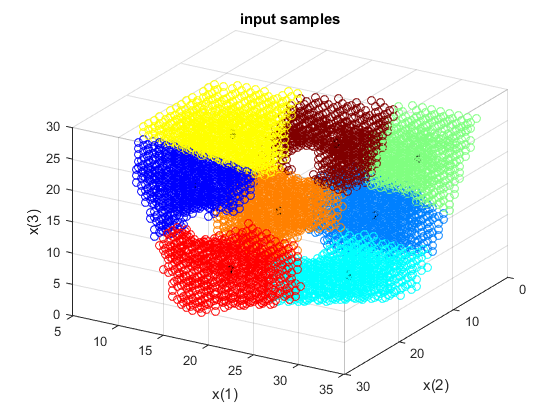}
 	\caption{Voronoi Tessellation for randomly chosen safe samples for the three obstacle scenario.The centroids of each region i.e. centers of the clusters are taken as milestones.In this illustration $K = 8$. }
 	\label{vt}
 \end{figure}
 \subsubsection{Building Probabilistic Road Map}\label{prm}
 \par{The subset of points chosen from target space are taken as vertices for the Probabilistic Road Map construction. K-Nearest Neighbours (KNN)\cite{c10} algorithm is used for determining the vertices adjacent to a given vertex and Euclidian distance is chosen as the comparison metric. The value of K has to be high enough to maximise the probability of having at least one kinematically feasible path when the graph is traversed in real-time but at the same time it is required to keep the value of K as moderate as possible to curtail the runtime complexity. We found that in a typical urban city scenario the value of K varies between 40 and 120 for optimal results.}
 \par{
 The initial order in which milestone way points are visited is determined from the nearest neighbour algorithm (greedy algorithm) which lets the UAV choose the nearest unvisited milestone as its next milestone.This technique has been chosen for approximating the shortest tour as it gives a path 25\% longer than the optimal path on average. Also, Tassiulas\cite{c11} has shown that a tight upper bound of \  $\frac{d\sqrt d}{d-1}\ast N^{\frac{d-1} d}+o(N^{\frac{d-1} d})$ exists for the nearest neighbour tour. Here N is the number of cities and d is the optimal path distance for the Euclidian travelling salesman problem wherein the triangle inequality holds.}
 \begin{figure}[!tbp]
 	\centering
 	\begin{subfigure}[b]{0.4\textwidth}
 		\includegraphics[width=\textwidth]{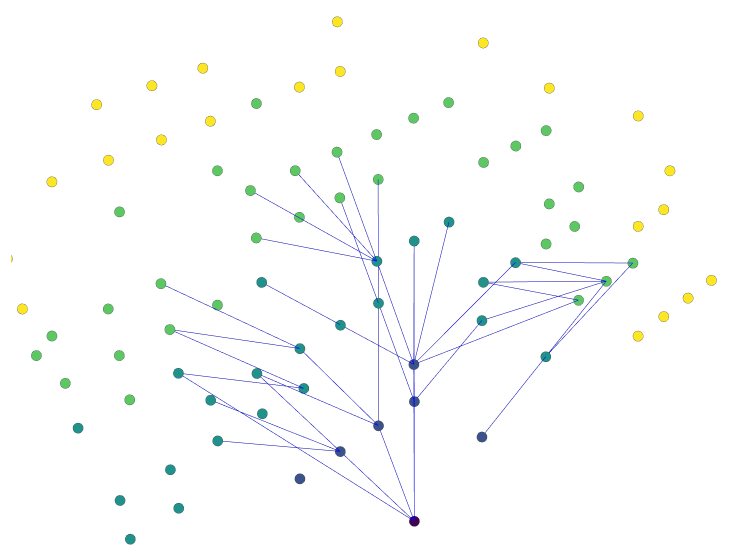}
 		\caption{Node Expansion from source node up to depth 3}
 	\end{subfigure}
 	\hfill
 	\begin{subfigure}[b]{0.4\textwidth}
 		\includegraphics[width=\textwidth]{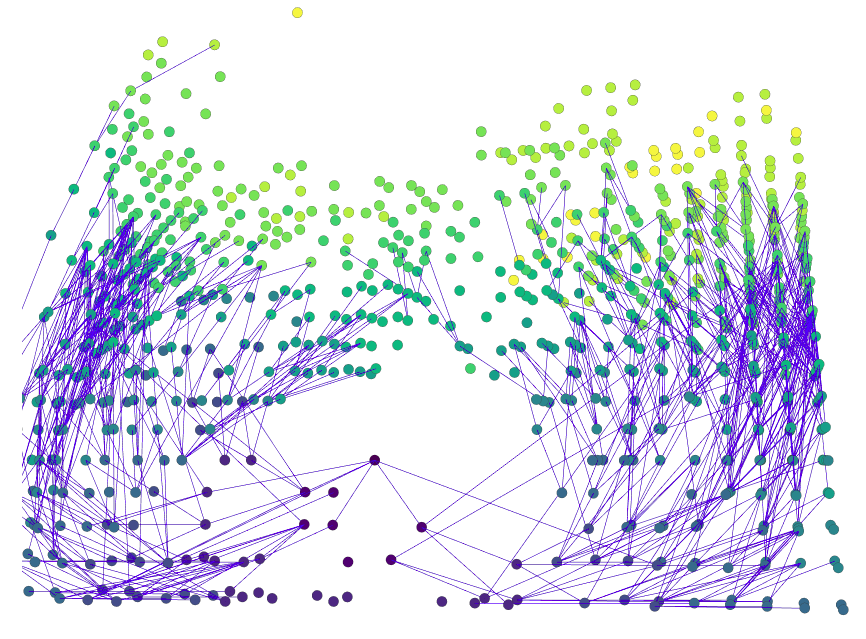}
 		\caption{Node Expansion from source node up to depth 10}
 	\end{subfigure}
 	\caption{PRM Node Expansion}
 \end{figure}
 \subsubsection{Kinematically feasible Path }
 Kinematic constraints of minimum turning radius  $R_{h,\mathit{min}}$ and maximum climb rate  $p_{max}$ \ for the UAV in current state  $(x,y,z,\omega ,\varphi ,\theta )$ are enforced during generation of motion primitives when the PRM is traversed using A* algorithm. At any given node, only those adjacent nodes are considered for shortest path
 which satisfies the kinematic constraints with respect to the current state of the UAV. This modification not only
 reduces the complexity of the graph but also leads to a smoother path for traversing. 
 
 Thus the node expansion from the current node depends not only on its distance from the current node and occupancy grid value but also on its spatial orientation from it.The heuristic chosen for selection of next best node from the list of adjacent nodes is given by
 \begin{equation}\label{astar}
       h^{(x,y,z)}=\ C^{(x,y,z)}_{Risk}+\ p^{(x,y,z)}_{revisitation}
 \end{equation}
 \begin{equation}
 p^{(x,y,z)}=Ke^{(V_{count}-1)}
 \end{equation}
 \begin{equation}
 	d^{(x,y,z)}={({(x_{curr}-x)}^2+{(y_{curr}-y)}^2+{(z_{curr}-z)}^2)}^{1/2}
 \end{equation}
 \begin{equation}
 	f^{(x,y,z)} = h^{(x,y,z)} + d^{(x,y,z)}
 \end{equation}
 where  $d$ is the Euclidian path cost,$V_{count}$ is the number of times the node has already been explored, $f$ is evaluation function for node expansion and $K=1000$ for imposing a high penality on revisitation. Revisitation penalty is imposed on an already visited node to
 ensure that preference is given to new unsearched areas rather than remaining confined to local minima and lose gatherable intelligence.
 \subsubsection{Pseudocode} 
 Algorithm 1 gives the pseudo code of the approach described in previous sections:
 \begin{algorithm}[h!]\label{algo1}
 	\SetKwInOut{Input}{Input}
 	\SetKwInOut{Output}{Output}
 	\Input{Geofence, Initial location and nature of Obstacles, UAV parameters,Source Point}
 	\Output{Gloabal Path waypoints}
 	$resolution\_distance\leftarrow2*turning\_radius$\;
 	\emph{$space\_matrix\leftarrow$get\_range$(resolution\_distance)$}\;
 	\emph{\textbf{Convert} Geodatic coordinates to ECEF}\;
 	\ForEach{point in space\_matrix }
 	{
 		\eIf{point is outside geofence or point is inside obstacle}
 		{
 			$point.score\leftarrow1$\;
 			
 		}
 		{
 			$global\_score(obstacle\_list,point)$\;
 			$local\_score(space\_matrix,point)$\;
 			$point.score \leftarrow max(global\_score,local\_score)$\;
 		}
 	}
 	sample\_list$\leftarrow$ Randomly Sample minimum required points $\forall$ points whose score $<\delta$\;
 	$milestones\leftarrow K-Medoids(sample\_list,iterations)$\;
 	$SetOrder(milestones,source\_point)$\;
 	initialize $PRM$\;
 	\ForEach{point in sample\_list}
 	{
 		inititalize node\;
 		Add node to PRM\;
 	}
 	\ForEach{node in PRM}
 	{
 		$neighbour\_list \leftarrow Nearest\_Neighbour(node,K)$\;
 		\ForEach{node in neighbour\_list}
 		{
 			\If{Collision\_Free(Extend(node,neighbour\_list))}
 			{$node.neighbour\_list \leftarrow node$\;}
 		}
 	}
 	initialize Global Path\;
 	\ForEach{node in milestone}
 	{
 		$path \leftarrow A^*(node,node \rightarrow next)$\;
 		$global\_path.append(path)$\;
 	}
 	
 	\caption{Global Path Generation}
 \end{algorithm}
\par{In lines 1-2 we discretize the search space based on resolution distance along all the three axes.In lines 4-12 collision risk score is computed for all the cells in the discretized grid.In line 13 random samples are acquired for which the collision risk score is less then the collision risk threshold($\delta$).In line 14 acquired samples are clustered and centroids of each cluster are set as milestones.In line 15, starting from the source point we rearrange the milestones according to the euclidean distance in a greedy manner such that nearest milestone to source point is taken first then the milestone nearest to the previous milestone which has not already been visited comes next till all milestones have been connected.In lines 16-28 PRM is built as described in Section \ref{prm}.In lines 29-33 the path segments between the consecutive milestones are computed using $A^{*}$ with heuristic given in Eq. \ref{astar}.Individual path segments are appended in order to obtain the final global path.}
\par{The computational complexities involved with each of the phases in the global planner are given as follows:
\begin{itemize}
	\item Initial occupancy grid initialisation and collision risk score assignment collectively have a time complexity of order of $O(KN^3)$.
	\item Voronoi tessellation of discretized search space on the randomly sampled safe points has an associated time complexity of order of $O(mnt)$.
	\item Probabilistic Roadmap building phase by means of K-Medoid Clustering with max heap optimisation runs in $O(n^2 logk)$. This phase along with $A^*$ run on chosen heuristic are the dominating factors during the execution of the algorithm.
	 
\end{itemize}
Here, K = number of obstacles, N=discrete division along one dimmension of search space, m= number of milestones/clusters, n= count of sampled safe points, t= number of iterations, k = count of nearest neighbours.
}

 \subsection{Local Planner}
 \noindent The path generated in the grid planning phase though feasible has its limitations. Continuously updating the global path by querying the grid planner in response to changes in the environment is a computationally expensive process and also the path generated by the global planner is jagged. To remedy these limitations and provide a higher resolution of control in motion planning a local planner is built atop the global planner.  The motion primitives generated between two configurations represented by successive nodes of the global path ensure that vehicle dynamics strictly adhere. 
 
 \noindent The local planner proposed generates motion primitives based on shortest Dubins path between the two configurations. The trajectory between these configurations is decomposed onto two orthogonal planes and for each of these 2D Dubins curves are calculated taking into account any obstacles that maybe present in between these configurations.
 \subsubsection{Dubins Motion Primitives}  
\begin{figure}[h!]
 	\centering
 	\includegraphics[width=\linewidth]{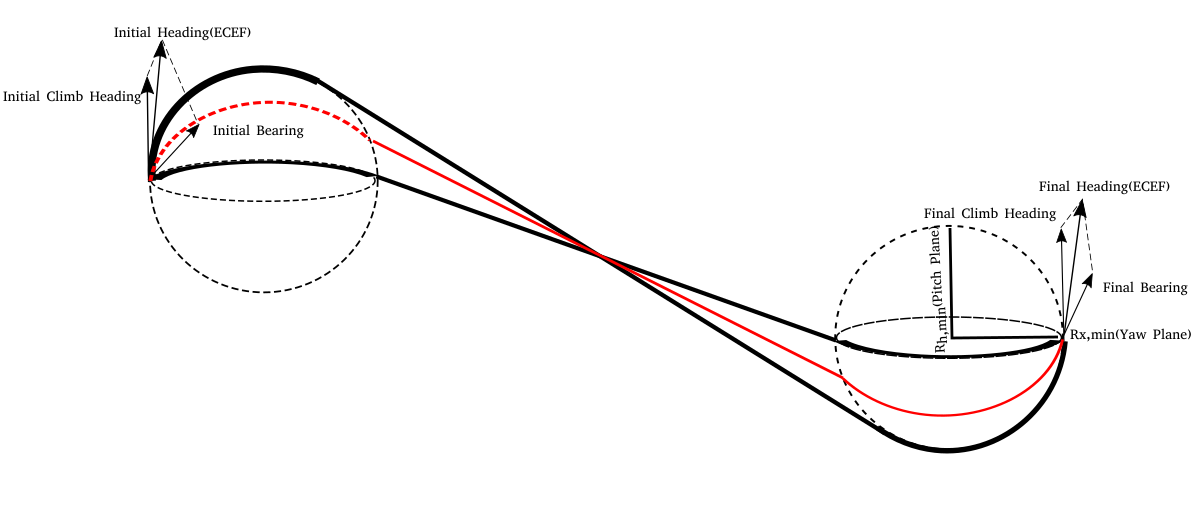}
 	\caption{3D Dubins Path}
 	\label{3ddubin}
 \end{figure}

 \noindent A node in the global path is defined as follows
 \begin{equation}
	 node=\left\{x,y,z,\psi,v,C_{risk}\ \right\}
 \end{equation} 
 where the first four components ($x,y,z,\psi)$ describe the location and desired heading vector, $v\ $represents the target airspeed and $C_{risk}$ is the collision risk score associated with the given node. Given any two consecutive nodes in the global path i.e.  ($x_1,y_1,z_1,\psi_1)\ \ $and ($x_2,y_2,z_2,\psi_2)$ the connecting path is constrained by minimum turning radius, initial and final bearings. The 3D path connecting the two nodes can be decomposed into two orthogonal planes as shown in Fig.\ref{3ddubin}.
 \par
\begin{figure}[h!]
   	\centering
   	\includegraphics[width=\linewidth]{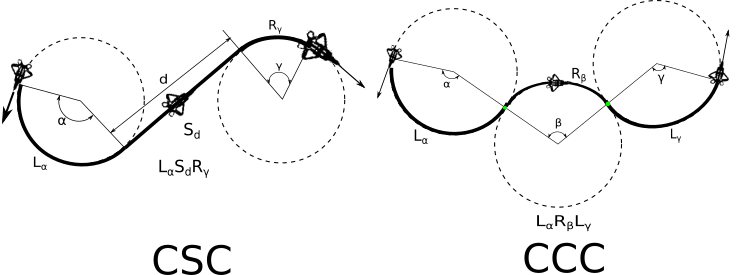}
   	\caption{2D Dubins Path}
   	\label{2ddubin}
   \end{figure}

{Bearing and minimum turning radius constrains UAV's movement in the horizontal plane while max climb rate and direction of climb(ascent/descent) affect UAV's propagation in the vertical plane. 2D Dubins\cite{c12} curve is created for each of these orthogonal planes. Depending on the distance of separation between the two nodes, either one of CSC or CCC trajectories[Fig.\ref{2ddubin}] is chosen for the shortest path. The benefit of choosing Dubins curves to join the nodes is twofold as the initial and final headings are automatically aligned and it renders the shortest path given the radius curvature constraints.}
  
\subsubsection{Obstacle Path Adjustment} 
 \noindent If an obstacle lies on the connecting path between the two consecutive nodes, then the Dubins path is modified to steer the UAV clear of obstruction with minimum deviation from the determined path. Depending on nature (static/dynamic) and radius of obstacle different path modification strategies are adopted. If the radius of the obstacle is less than the minimum turning radius of UAV, then its radius is taken to be equal to $R_{min}$ for computation of viable curves in resultant Dubins path. 
 \paragraph{Static Obstacle}
 For a static obstacle multiple paths exist that can be generated for safe propagation of UAV, to ensure minimum deviation, path passing through tangentially closest point on the periphery of the obstacle is chosen, given that the change in steering angle for such a point is feasible.Assume that the UAV is initially at $x_{i},y_{i},z_{i}$ and going towards next node at $x_{f},y_{f},z_{f}$,then the initial bearing and heading of the UAV is given as $\psi_{initial}$ and $\theta_{initial}$ where
 \begin{equation}
 	 \psi_{iniital} = \frac{\pi}{2}- {{\mathrm{tan}}^{-1} (\frac{y_f-y_i}{x_f-x_i})\ }
 \end{equation}
 \begin{equation}
 {\theta }_{init}=\ {{\mathrm{tan}}^{-1} (\frac{z_f-z_i}{y_f-y_i})\ }
 \end{equation}
 Also consider a static obstacle that appears at $x_{o},y_{o},z_{o}$ with radius $R_{obst}$ and height $h_{obst}$.The obstacle can be dodged in two ways either by changing the current yaw or pitch angle.
 \par{The change in yaw angle($\Delta\psi_{yaw}$) is given as}
 \begin{equation}\label{yaw}
 \Delta \psi_{yaw}=\ \psi_{final} -\ \psi_{inital}
 \end{equation}
 where $\psi_{final}$ is expressed as
 \begin{equation}\label{psifinal}
 \psi_{final} =\ {{\mathrm{cot}}^{-1} (\frac{y_{dodge}-y_i}{x_{dodge}-x_i})\ }
 \end{equation}
 here $x_{dodge},y_{dodge}$ are x-y coordinates of the tangentially closest point on the periphery of the obstacle required to dodge the obstacle and can be given as
 \begin{equation}\label{xdodge}
 	x_{dodge}=\ x_0+\ R_{buff}{\mathrm{cos} ({{\mathrm{tan}}^{-1} (m_p)\ })\ }
 \end{equation}
 \begin{equation}\label{ydodge}
 	y_{dodge}=\ y_0+\ R_{buff}{\mathrm{sin} ({{\mathrm{tan}}^{-1} (m_p)\ })\ }
 \end{equation}
 where \begin{equation}\label{buff}
 	R_{buff} = R_{obst} + d_{buff}
 \end{equation}
 and
 \begin{equation}\label{mp}
 	m_p = \frac{(x_i - x_o)^2 - R_{buff}^2}{(x_i - x_o)(y_i - y_o) \pm R_{buff}((l^2 - R_{buff}^2)^{1/2} }
 \end{equation}
In Eq.\ref{buff}  $d_{buff}$ is minimum safety clearance. $l$ denotes the euclidean distance between $(x_o,y_o)$ and $(x_i,y_i)$. The z-axis is taken as the axis of rotation for yaw and hence there is no change in z coordinate.
\par{Similarly the change in pitch angle is given as 
	}
 \begin{equation}\label{thetafinal}
 	\Delta \theta_{pitch} =\ {\theta }_{final}-{\theta }_{initial}
 \end{equation}
 where ${\theta }_{fin}$ is expressed as
 \begin{equation}
 {\theta }_{fin}=\ {{\mathrm{tan}}^{-1} (\frac{z_{dodge}-z_i}{y_{dodge}-y_i})\ }
 \end{equation} 
 here $z_{dodge},y_{dodge}$ are y-z coordinates of the tangentially closest point on the periphery of the obstacle required to dodge the obstacle and can be given as
 \begin{equation}
 	z_{dodge}=\ h_{obst}+\ d_{buff}
 \end{equation}
 \begin{equation}
 	y_{dodge}=\ y_0-\ R_{buff}
 \end{equation}
 The x-axis is taken as the axis of rotation for pitch and hence there is no change in x coordinate.The appropriate way to dodge the obstacle by changing pitch ot yaw is determined by wheather $\Delta\psi_{yaw}$ and $\Delta\theta_{pitch}$ are within UAV's constraints. If both are feasible the smaller change in angle requirement is followed. 
 \paragraph{Dynamic Obstacle}   
 Consider the inital and target nodes as given in the previous section.Also consider a dynamic obstacle that appears at $x_{o},y_{o},z_{o}$ and velocity $\overrightarrow{V}$.A similar procedure as followed to dodge the static obstacle is followed here with the difference being that $x_{o},y_{o},z_{o}$ are not fixed and hence a sequence of dodge points are generated until the UAV steer clears of the obstacle.
 \par{Again the UAV can dodge by either changing the yaw or pitch to generate the next dodge point.At any instant the computation for the next dodge point as the result of the change in yaw angle is similar to that of the static obstacle and is given by Eq. \ref{yaw},\ref{psifinal},\ref{xdodge},\ref{ydodge}.}
 \par{The change required in pitch angle(heading) is given as in Eq. \ref{thetafinal} but here 
 	}
 	 \begin{equation}
 	 \theta_{k}= \tan^{-1}(\frac{z^{(k)}_{dodge} - z^{(k-1)}_{dodge}}{y^{(k)}_{dodge} - y^{(k-1)}_{dodge}})
 	 \end{equation} 
 	 where
 \begin{equation}
 z_{dodge}^{(k)}=\ z_{o}+\ R_{buff}{\mathrm{sin} ({{\mathrm{tan}}^{-1} (m_q)\ })\ }
 \end{equation}
 \begin{equation}
 y_{dodge}^{(k)}=\ y_{o}+\ R_{buff}{\mathrm{cos} ({{\mathrm{tan}}^{-1} (m_q)\ })\ }
 \end{equation}
 and 
 \begin{equation}\label{mq}
 	m_q = \frac{(y_{dodge}^{(k-1)} - y_o)^2 - R_{buff}^2}{(y_{dogde}^{(k-1)} - y_o)(z_{dodge}^{(k-1)} - z_o) \pm R_{buff}(l^2 - R_{buff}^2)^{1/2} }
 \end{equation}
Here $(y,z)^{(k)}_{dodge}$ is the $k^{th}$ dodge point in the sequence,$(y,z)^{(k-1)}_{dodge}$ is the previous dodge point in the y-z plane and l is the euclidean distance between $(y_o,z_o)$ and $(y_{dodge}^{(k-1)},z_{dodge}^{k-1})$.The dodge point generation is terminated when the bearing from the current dodge point to the target point($x_{f},y_{f},z_{f}$) is no longer intercepted by the obstacle.For more details on $m_p$ and $m_q$ refer to Appendix.
 		
 		\subsubsection{L1 Navigation Controller}
 		\noindent A non-linear navigation controller based on\cite{c13} was used to guide the UAV along curved segments of the trajectory. The controller not only works like a PD controller for straight paths but also contains an anticipatory control element for minimizing drifts when following curved paths. The guidance logic works by selecting a reference point at a distance $L_1$ on the desired trajectory in front of the UAV. Then a lateral acceleration command is generated which gives it an adaptive capability to face external disturbances like winds. The expression for lateral acceleration command is given by
 		\begin{equation}
 			a_{cmd}=2V^2{\mathrm{sin} \tau /L_1}
 		\end{equation}
 		 
 		Where $V$ is the velocity of UAV, $L_1$is defined from current position to reference point, $\tau $ is the angle between $V$ and${\ L}_1$.  Details regarding the calculation of the expression and error estimation are omitted due to space constraints and can be found in \cite{c13}.

\bigskip
\subsection{Markov Decision Process}
\noindent Three sets of actions can be adopted to prevent a collision with obstacles. A control policy is needed for determining the efficacy of each action in a given state. The control policy determines whether variation in velocity, local path adjustment, global path amendment or a combination is required to evade an impending obstacle or tackle changes in the environment.  The UAV control problem and decision process is modeled as a finite state Markov Decision Process\cite{c14}.

\begin{figure}[h!]
	\centering
	\includegraphics[width=0.75\linewidth]{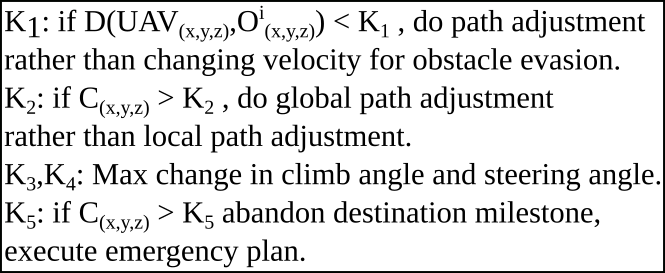}
	\caption{Parameters learned by Desicion Process}
	\label{mdp}
\end{figure}

\noindent We made use of the Policy Search algorithm as given in [15] to learn the critical parameters of our control policy.  For the reasons of brevity, only a short description of each of the learned parameters is given in Fig. \ref{mdp}
\section{Results and Experiments}

\par{We validate our algorithm on the sig rascal model provided by JSBSim as a simulation platform for a fixed wing UAV. The on-board autopilot, control system and sensors are simulated using the ArduPilot platform with APM Planner as the ground control station. A custom C\# .Net application was built for controlling and configuring the obstacles and Caesium was used for depicting the final scenarios in real-time. The kinematic constraints for the considered system are given in Table \ref{tab:table1}.}
\begin{table}[h!]
	\centering
	\caption{Parameters for Simulation}
	\label{tab:table1}
	\begin{tabular}{ll}
		\toprule
		Parameters & Values\\
		\midrule
		Turning Radius & $22m$\\
		Climb-Rate & $8ms^{-1}$\\
		Cruise Speed & $30ms^{-1}$\\
		Max. Climb-Angle & $30^{\circ}$\\
		Max. Roll Angle & $65^{\circ}$\\
		Max. Pitch Angle & $25^{\circ}$\\
		Min. Pitch Angle & $-20^{\circ}$\\
		NAV L1 Damping & $0.75$\\
		NAV L1 Period & $15.0$\\
		NAV L1 X TRAC I & $0.2$\\
		\bottomrule
	\end{tabular}
\end{table}
\par{We conducted over 44 full system simulations on a workstation equipped with an octa core intel i7, 2.50 Ghz CPU.In any typical mission there were 3-6 stationary obstacles of varying radii spread across the search space and 1-3 dynamic obstacles traversing along predetermined paths. The criterion involved in evaluating our system were the fraction of total search area (enclosed within the geofence boundary) surveyed, length of the path and distance of separation of UAV from the obstacles within the search space.}
\begin{figure}[h!]
\centering
\includegraphics[width=0.95\linewidth]{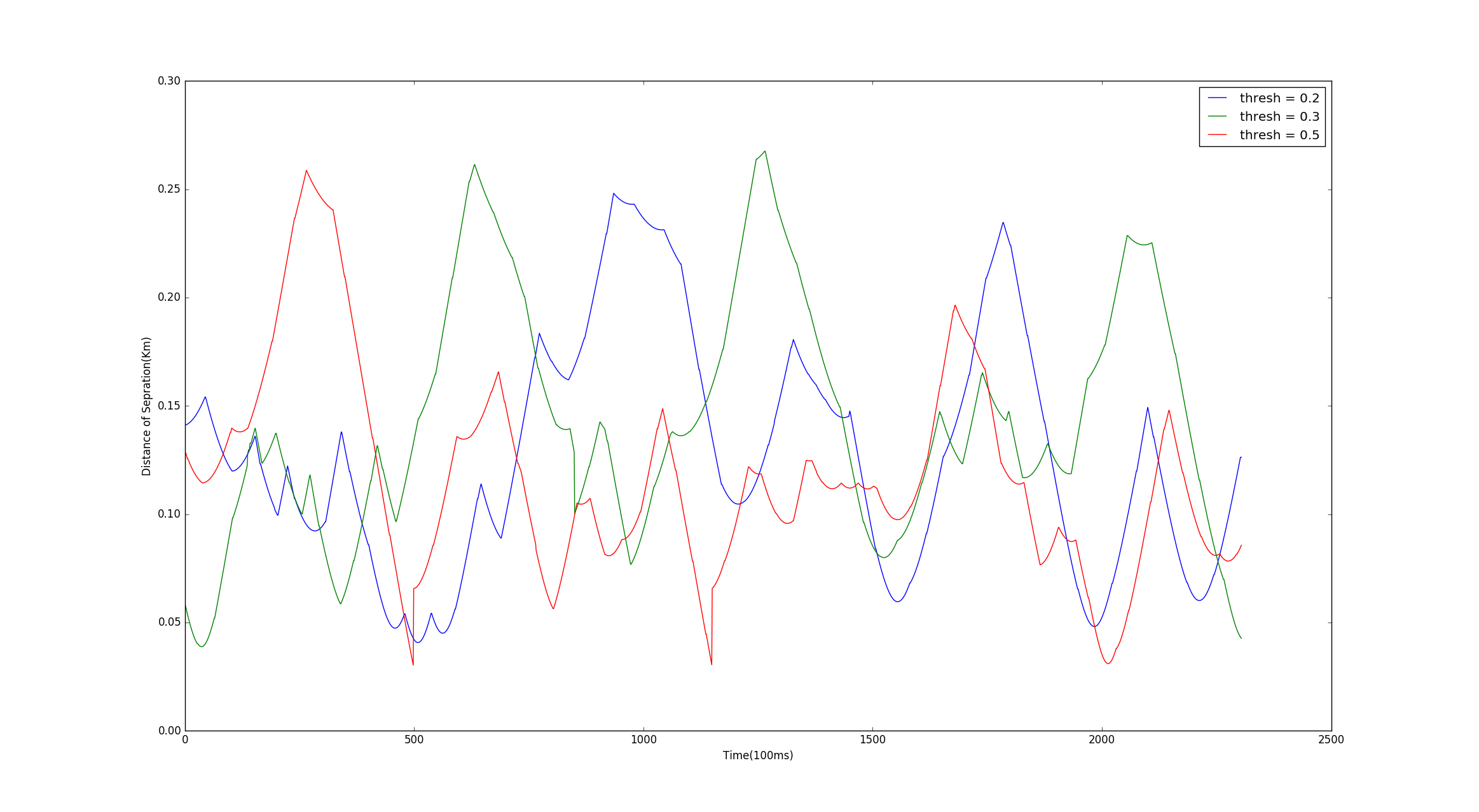}
\caption{Shows the average distance of separation of UAV from the obstacles during the course of the mission for different values of collision risk threshold with the number of milestones fixed at 10. }
\label{sepvstime}
\end{figure}
\begin{figure}[h!]
	\centering
	\includegraphics[width=0.95\linewidth]{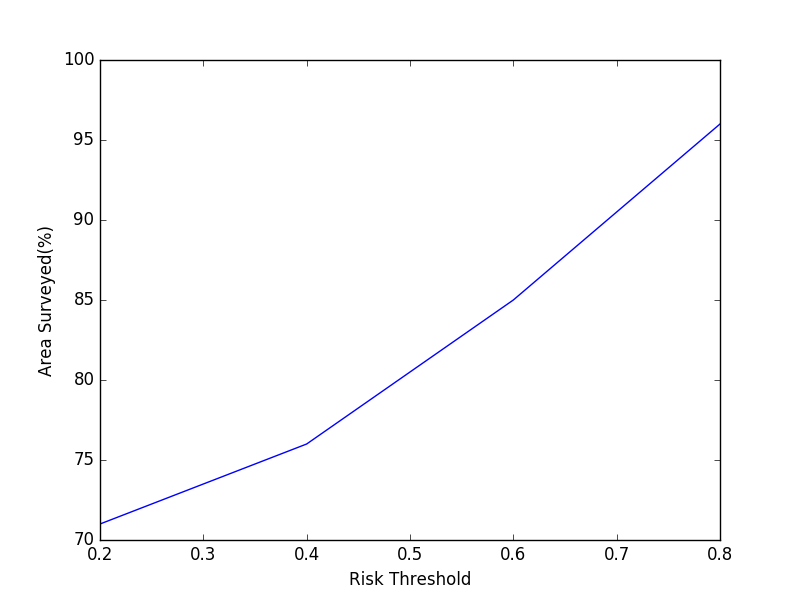}
	\caption{Area Surveyed vs Collision Risk Threshold}
	\label{risk}
	
\end{figure}
\par{
From Fig.\ref{sepvstime} and Fig.\ref{risk} we conclude that the minimum distance of separation of UAV from  obstacles decreases as collision risk threshold is increased from 0.2 to 0.8 at the same time the surveyed region increases from 72\% to 94.7\%.Hence an appropriate collision risk threshold has to be estimated experimentally for the desired UAV.Ideally the value lies between 0.4 - 0.7 depending on the mission profile.}

\begin{figure}[h!]
	\centering
	\includegraphics[width=0.48\linewidth]{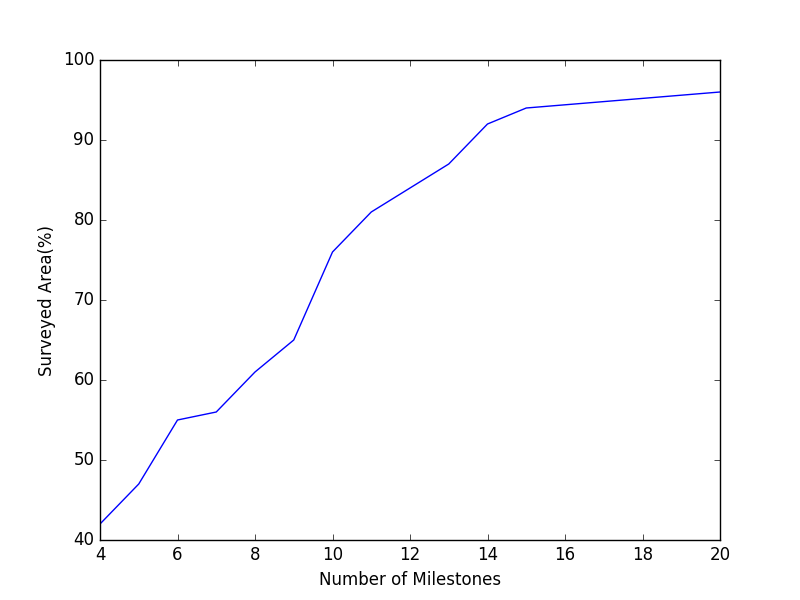}\hfill
	\includegraphics[width=0.48\linewidth]{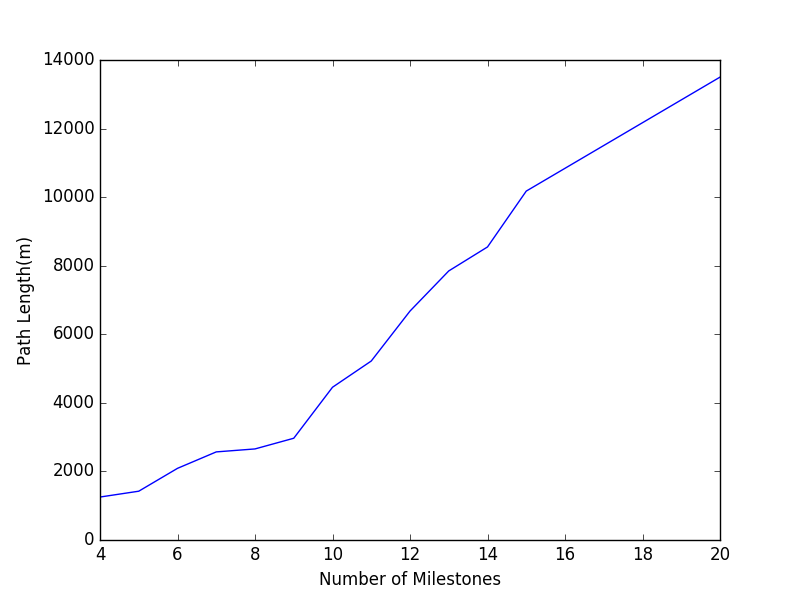}
	\caption{(a)Area Surveyed vs Number of Milestones (b)Path Length vs Number of Milestones}
	\label{mile}
\end{figure}
\begin{figure}[h!]
	\centering
	\includegraphics[width=0.95\linewidth]{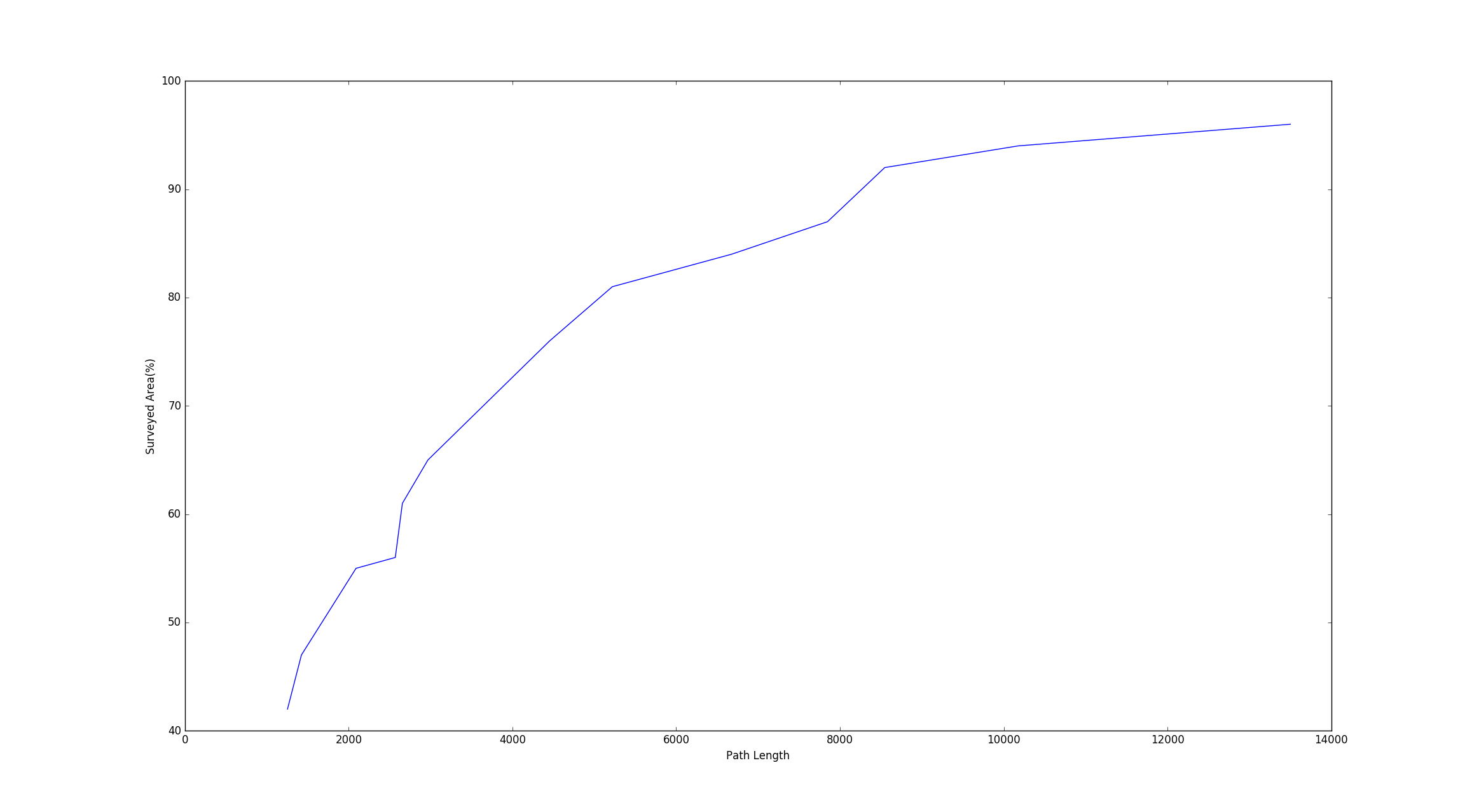}
	\caption{Trade-off between Area Surveyed and Path Length for different Milestone count}
	\label{conc}
\end{figure}
\par{The choice of number of milestone waypoints for mission also affects the expanse of search space surveyed. Fig. \ref{mile}(a) shows the increase in the area covered with increase in number of milestones. From Fig.\ref{mile}(b) it is also noticed that the length of the tour also increases substantially.From Fig.\ref{conc} it is noticed that there exists a trade-off between area surveyed and path length with respect to milestone count.  In our simulations we found that for the geofence considered, there is a consistent increase in area surveyed with gradual increase in path length till 14 milestones, beyond which the path length increases without any appreciable growth in area surveyed. This is consistent with our heuristic as proposed in Eq.\ref{fov}  where for an area of dimensions 834x577$m^2$ the optimal milestone count should be 14 for a mission in which the on-board camera has fov $60^o$ and anticipated average altitude is 180$m$.} 
\begin{table}[h!]
	\centering
	\caption{Computation SpeedUp}
	\label{tab:table2}
	\begin{tabular}{cccc}
		\toprule
		C-Space Dim. & A* & A*-PRM & Speedup\\
		\midrule
		25x30x27 & 27685 & 3109 & 8.90\\
		40x30x30 & 31084 & 5942 & 5.23\\
		50x50x50 & 52031 & 6232 & 8.95\\		
		\bottomrule
	\end{tabular}
\end{table}
\par{Precomputation of a probabilistic roadmap from sample space greatly reduces planning time. Each node in PRM is extended by connecting to K-Nearest Neighbours and pruning those extensions which do not adhere to the dynamic and kinematic constraints of the system, then A* is used to find the optimal path between the milestone nodes in PRM. In comparison applying A* over the complete search space takes much longer to find a similar path with no added benefit. Table \ref{tab:table2} shows the speedup in computation when a precomputed PRM is used instead of searching through the entire search space. On an average, the global path planning phase runs 7.69 times faster due to this optimization.}  
\begin{figure}[h!]
	\centering
	\includegraphics[width=0.48\linewidth]{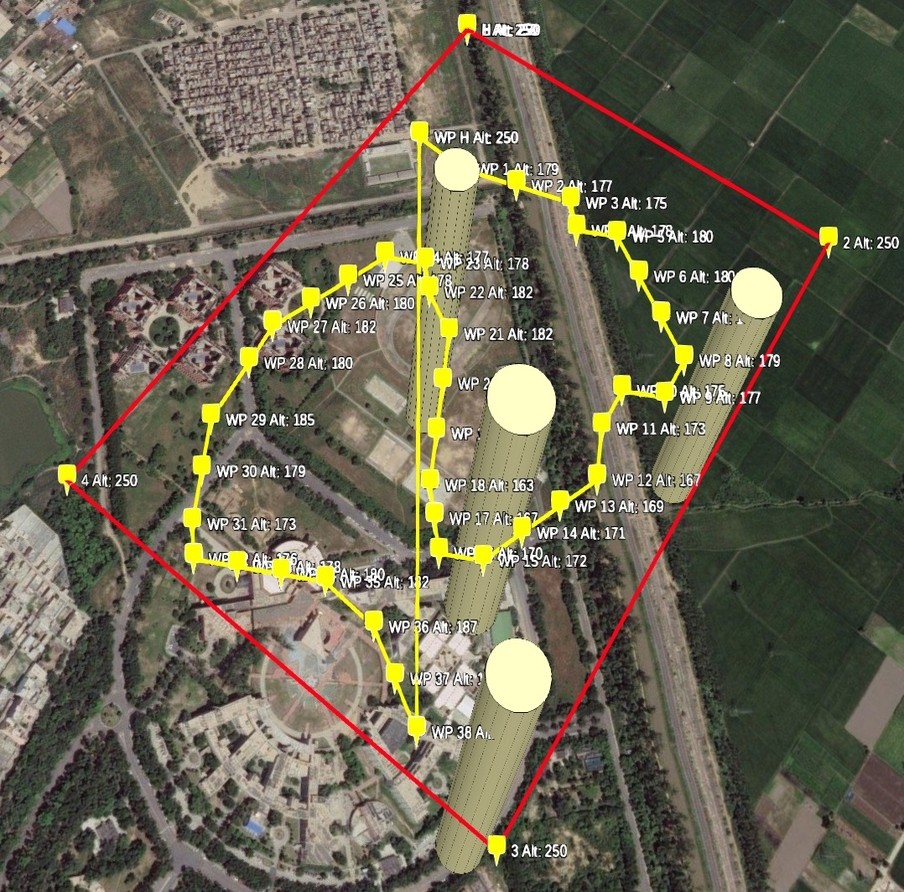}
	\includegraphics[width=0.48\linewidth]{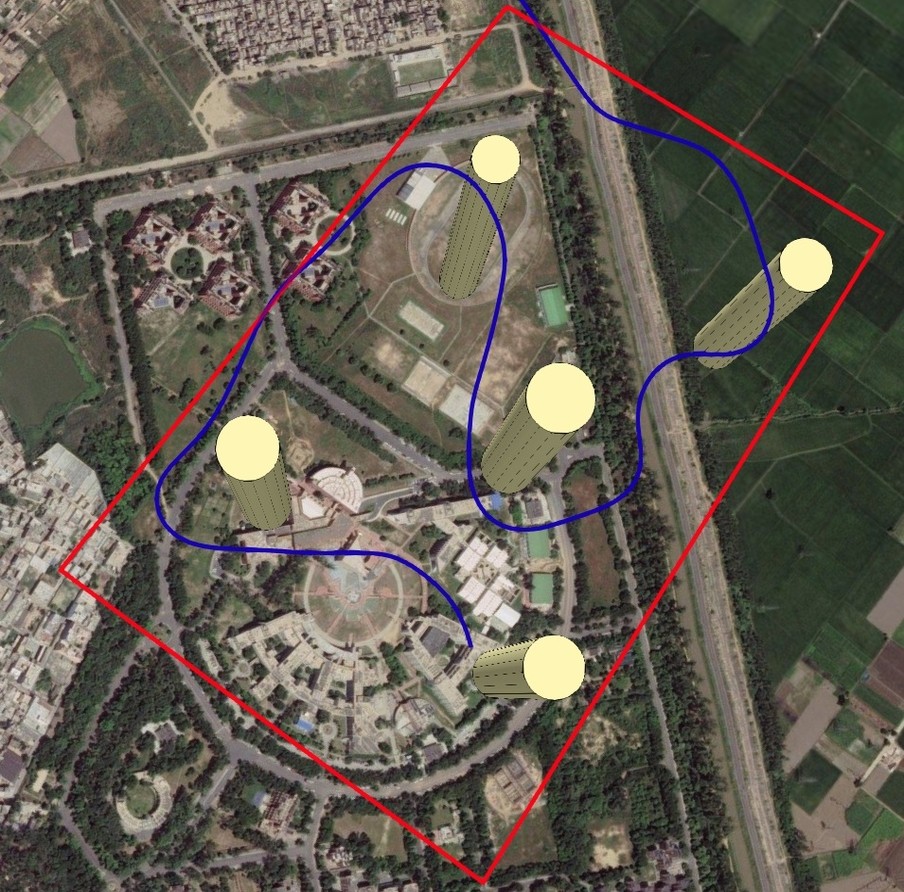}
	\caption{(a)Red boundary represents the Geofence,Yellow curve depicts Initial Global Path (b)Blue curve is final Smooth Trajectory of the UAV followed in Simulation after Dubins smoothing}
	\label{basic}
\end{figure}
\begin{figure}
	\includegraphics[width=.48\linewidth]{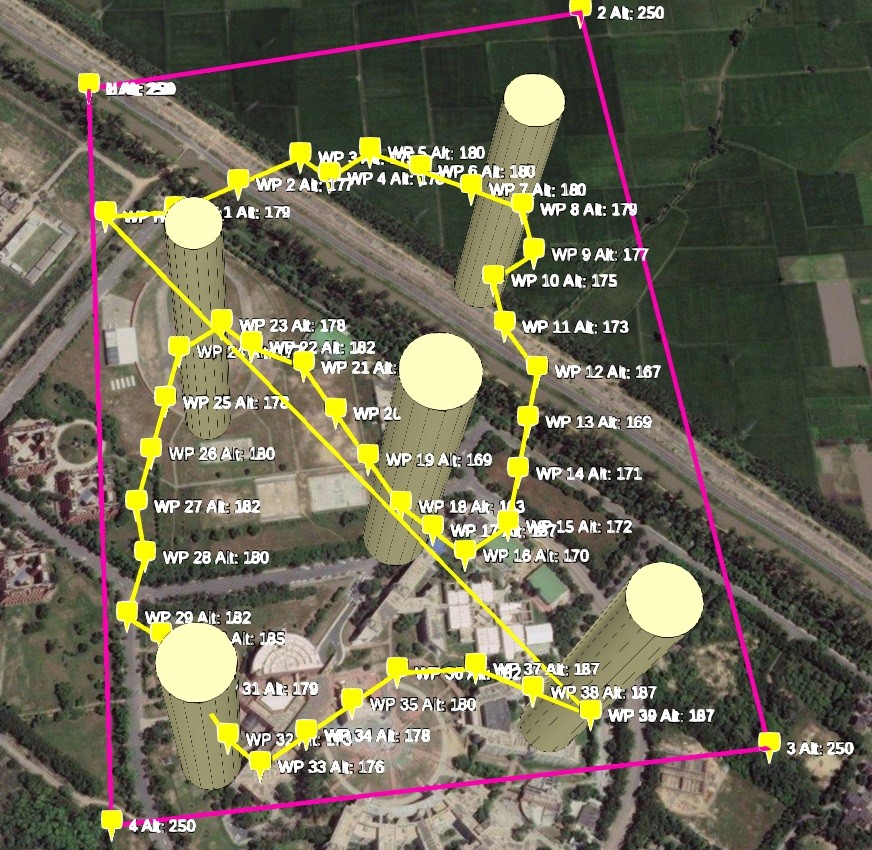}\hfill
	\includegraphics[width=.48\linewidth]{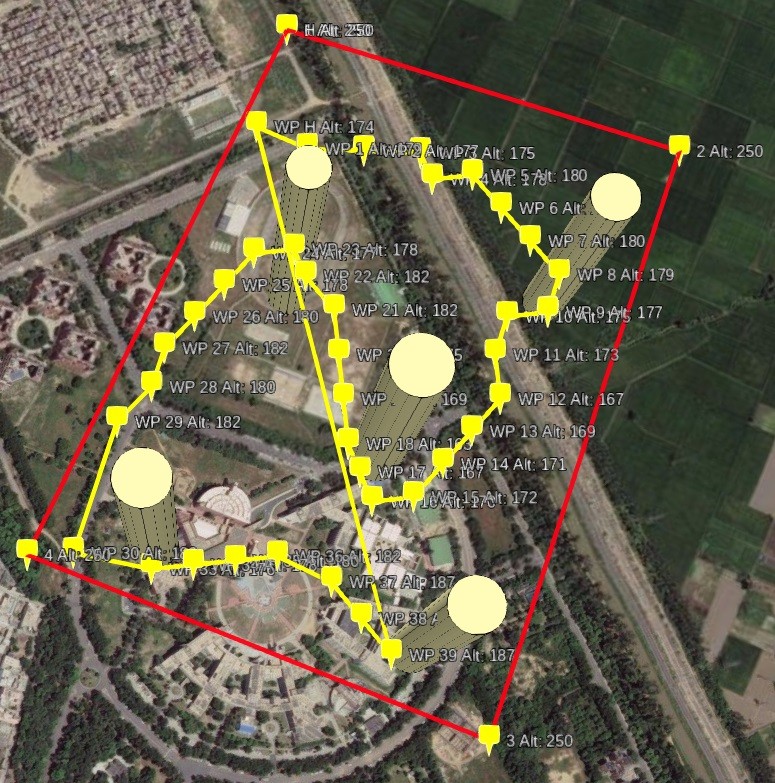}
	\caption{(a)Fifth Obstacle appears intersecting the path of the UAV (b)Corrected path in response to the new obstacle}
	\label{static}
\end{figure}

\begin{figure}
	\includegraphics[width=.48\linewidth]{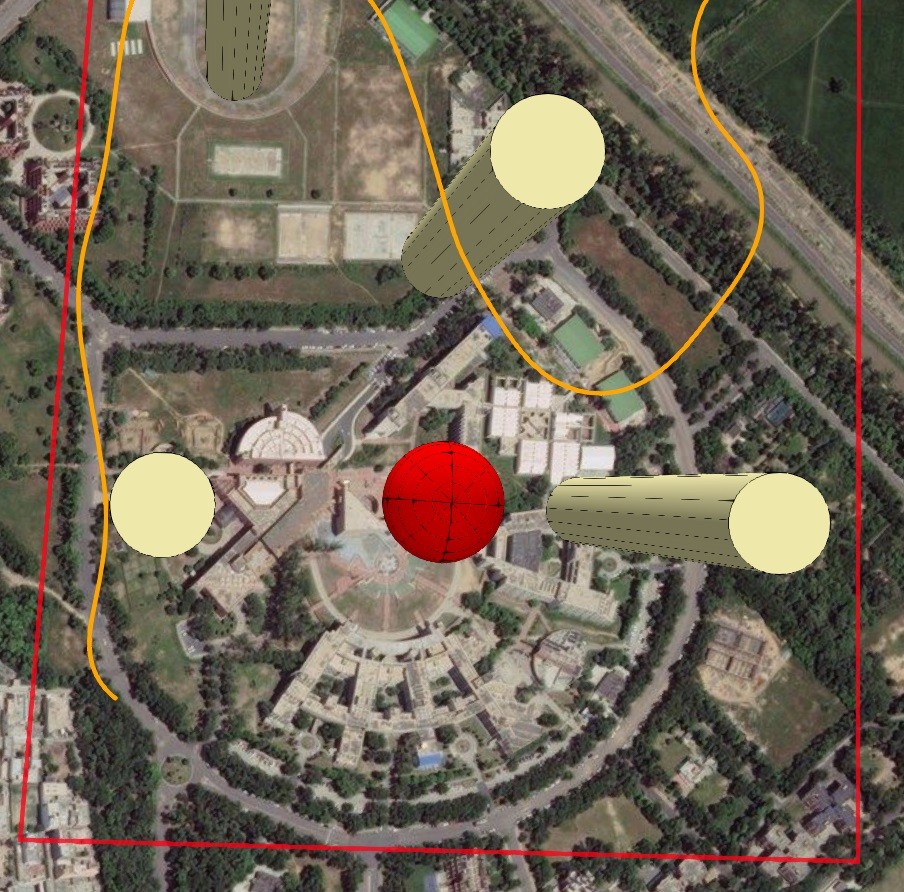}\hfill
	\includegraphics[width=.48\linewidth]{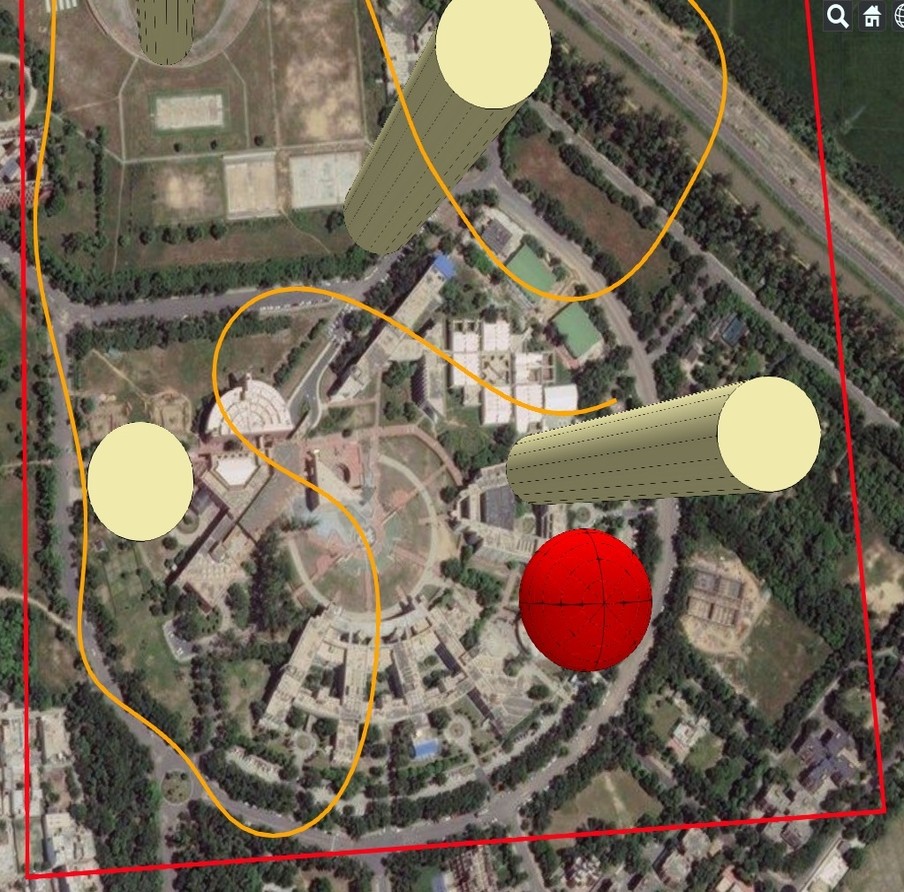}
	\caption{(a)UAV detects the approaching dynamic obstacle(b)Final Path taken by the UAV:A sequence of dodge points is added tangentially by the local planner to evade the obstacle}
	\label{dynamic}
\end{figure}
\noindent 
Fig.\ref{basic}, \ref{static}and \ref{dynamic} illustrate the results of simulations, in which a consistent wind blowing at 2 m/s with the bearing of 120$\mathrm{\textrm{°}}$ from North was considered. Non-linear L1 controller with the aforementioned parameters was used as the waypoint controller to ensure that the UAV does not drift considerably from its trajectory in presence of wind. For all the illustrations the boundary of geofence was demarcated at points A (28.754812 N, 77.115204 E), B (28.754812 N, 77.115204 E), C (28.755188 N, 77.119689 E) and D (28.748698 N,77.120161 E). The UAV after take-off enters the search space at coordinates at point (28.752088 N,77.116211 E). Fig.\ref{basic} demonstrates the working of the global grid planner in the presence of obstacles , four stationary obstacles of radii 25,35,28 and 30 $m$ are present at locations (28.7536640 N,77.1160412 E), (28.7522719 N,77.1180367 E),(28.7547551,77.1193027) and (28.7503155 N,77.1195173 E). The global planner generates a kinematically feasible path represented by the yellow coarse path as shown in Fig.\ref{basic}(a).The local planner generates inter-connecting motion primitives which result in a smooth dynamically feasible path which is traversed by the UAV as shown in Fig.\ref{basic}(b).

Fig.\ref{static} depicts the response of the local planner when an additional fifth obstacle appears unexpectedly at (28.7502026 N,77.1157837 E) of radius 32$m$.The corrected path is shown in Fig.\ref{static}(b).We see that an additional dodde point has been added to the current path so as to avert the collision. 

Fig.\ref{dynamic} illustrates the case where in addition to previous obstacles a dynamic obstacle of radius 30$m$ approaches the path of the UAV with velocity 15$m/s$ at distance of 105$m$.A path alteration was made by the local planner to steer the UAV away safely from the oncoming obstacle. The UAV was able to stay clear of the obstacles and maintain a minimum separation of more than 3$m$ which was taken as the safety clearance.
\begin{figure}[h!]
	\centering
	\includegraphics[width=0.75\linewidth]{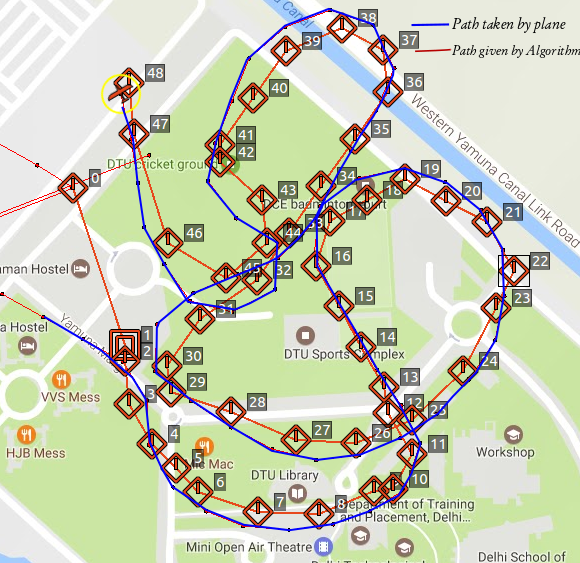}
	\caption{The orange curve represents global path, blue curve depicts actual path followed during simulation in APM Planner.}
	\label{sitll}
\end{figure}
 

\section{Conclusion}
In this study, a dynamic motion planning approach has been presented for a fixed wing UAV surveying urban environment. The specialty of the proposed approach lies in its utility to work in partially known and updating environment.The computation involved in generating feasible trajectories for the UAV through the search space is greatly reduced by the use of precomputed PRM.Voronoi Tessellation of search space and identification of key waypoints in the form of milestones leads to efficient mapping of the region to be surveyed.The changes in the environment are handled effectively by the decision process in the form of local or global planner.The application of 3D Dubins curve leads to smooth and dynamically feasible trajectories at runtime.The efficacy of the proposed approach is demonstrated in the complex simulation scenarios in highly cluttered environments comprising of both static and dynamic elements.In future, we plan on extending the capability of the proposed system by integrating an array of sensors to impart perception ability to our UAV platform.

\appendix
\section{Derivation for dodge point coordinates}
Derivation of dodge point coordinates for Eq. \ref{mp} and \ref{mq}.
\begin{figure}[h!]
	\centering
	\includegraphics[width=0.75\linewidth,height=0.75\linewidth]{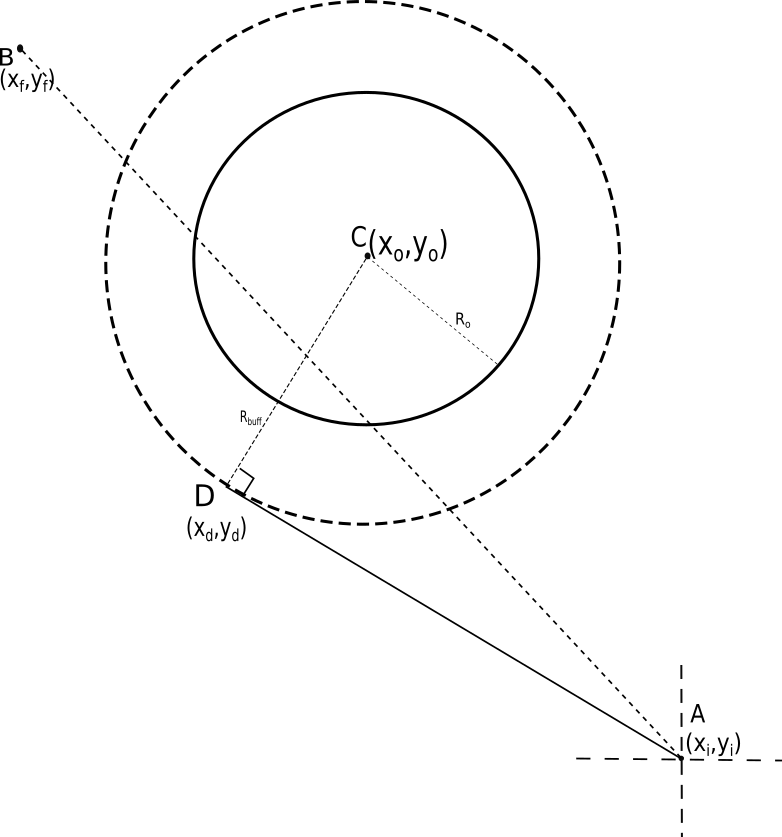}
	\caption{A typical dodging scenario}
	\label{derivation}
\end{figure}

Equation of the circle centered at $(x_o,y_o)$ is given as:
\begin{equation}\label{solve}
	(x - x_o)^2 + (y - y_o)^2 = R_{buff}^2
\end{equation}
Let the Equation of the tangent from point A be of the form
\begin{equation}
	y = mx + c
\end{equation}
then,after substituting $y$ and $c$ Eq. \ref{solve} can be written as:
\begin{equation}\label{solve2}
	(x - x_o)^2 + (mx + y_i - mx_i - y_o)^2 = R_{buff}^2
\end{equation} 
Since line AD is a tangent to the circle, therefore both the roots of Eq. \ref{solve2} are equal and hence the discriminant is zero. Using this we obtain the following quadratic relation in m:
\begin{equation}\label{quad}
	m^2(R{buff}^2 - X^2) - 2m(Y)(X) - (Y)^2 + R_{buff}^2 = 0
\end{equation}
where
	\begin{equation}
		X = x_i - x_o
	\end{equation}
	\begin{equation}
		Y = y_i -y_o
	\end{equation}
Solving the quadratic Eq. \ref{quad} we get the roots as:
\begin{equation}
	m = \frac{XY \pm R_{buff}(X^2 + Y^2 - R_{buff)^2})^{1/2}}{R_{buff}^2 - X^2 }
\end{equation}
where m is the possible values of the slope of line AD. The slope of line CD which is perpendicular to line AD is given by $-1/m$. Thus coordinates of the dodge points in the yaw plane are obtained as in Eq. \ref{xdodge} and \ref{ydodge} using parametric form equation of a line. Similar steps can be followed for pitch plane to obtain coordinates of dodge points.

\end{document}